\documentclass[letterpaper, 10 pt, conference]{ieeeconf}  

\usepackage{float}
\usepackage{cite}
\usepackage{graphicx}
\usepackage{booktabs}
\usepackage{multirow}
\usepackage{amsmath}
\usepackage{amssymb}

\IEEEoverridecommandlockouts                              

\title{\LARGE \bf
Learning to Score Sign Language with Two-stage Method
}

\author{ Hongli Wen$^{1}$ ,Yang Xu$^{2}$
\thanks{$^{1}$ Hongli Wen is with School of Artificial Intelligence, Beijing Normal University, Beijing 100088 
, China. 202111150036@mail.bun.edu.cn}
\thanks{$^{2}$ Yang Xu is with School of Artificial Intelligence, Beijing Normal University, Beijing 100088
, China. 202111081068@mail.bun.edu.cn}
}

\begin{document}

\maketitle
\thispagestyle{empty}
\pagestyle{empty}

\begin{abstract}
Human action recognition and performance assessment have been hot research topics in recent years. Recognition problems have mature solutions in the field of sign language, but past research in performance analysis has focused on competitive sports and medical training, overlooking the scoring assessment ,which is an important part of sign language teaching digitalization. In this paper, we analyze the existing technologies for performance assessment and adopt methods that perform well in human pose reconstruction tasks combined with motion rotation embedded expressions, proposing a two-stage sign language performance evaluation pipeline. Our analysis shows that choosing reconstruction tasks in the first stage can provide more expressive features, and using smoothing methods can provide an effective reference for assessment. Experiments show that our method provides good score feedback mechanisms and high consistency with professional assessments compared to end-to-end evaluations.
\end{abstract}

\section{Introduction}
China has approximately 85 million people with disabilities, among which more than 21 million have hearing and speech impairments, representing a significant demographic. However, as the sole communication language for the deaf and mute community, sign language faces numerous challenges in its learning and dissemination. Establishing a sign language teaching system and achieving the digitalization of sign language instruction will become an important part of computer-assisted motion training. Researchers worldwide have contributed to this topic, and some effective automated sign language teaching systems have already been proposed \cite{ref9,ref10,ref11}. Yet, in these methods using 3D human figures, there is usually only unidirectional input from learners, and it is almost impossible to find effective feedback interaction mechanisms, which contradicts the general principles of teaching. Traditional teaching experience shows that feedback on one's actions along with quantified scores are necessary.

Performance scoring is common in competitive sports, where one of the objectives is to achieve higher scores. Excellent research in this field \cite{ref12,ref13,ref14} often employs single-stage solutions, where end-to-end methods directly predict scores from videos. This is also due to the fixed camera angles, single video inputs, and rich quantitative annotations typical of these approaches. However, recent research oriented towards sports coaching \cite{ref15,ref16} has gradually introduced two-stage methods in flexible scenarios. These methods extract interpretable motion information, such as human skeletons or Mesh, and combine it with reference videos for evaluation. Prior knowledge is provided in the form of reference videos instead of simple labels, presenting significant challenges that require careful handling of time and space.

Some sign language teaching systems with feedback mechanisms have been proposed \cite{ref17,ref18}, but these methods typically use classification tasks to determine simple right or wrong responses, or directly regress scores. Similar to the first approach, sign language, as a communicative skill, should be assessed from the perspective of the latter approach. To this end, we propose a two-stage 3D sign language evaluation method, which takes a monocular sign language video, reconstructs human hand features in the first stage, and then in the second stage, smooths, embeds, and aligns the feature sequences with standard reference actions, ultimately producing scores that are educationally meaningful and highly consistent with expert evaluations. We hope this work will provide a reference for the design of feedback mechanisms in the development of digital sign language teaching systems.

The rest of this paper is organized as follows: Section II introduces related work on monocular human reconstruction, motion evaluation, and sign language motion evaluation. Section III discusses the detailed design and training process of the two stages of our system. Section IV validates the consistency of our method's scores with expert evaluations and user feedback. Finally, the paper summarizes work on incremental data extraction and looks forward to future research.

\begin{figure*}[h] 
\centering 
\includegraphics[width=0.9\textwidth]{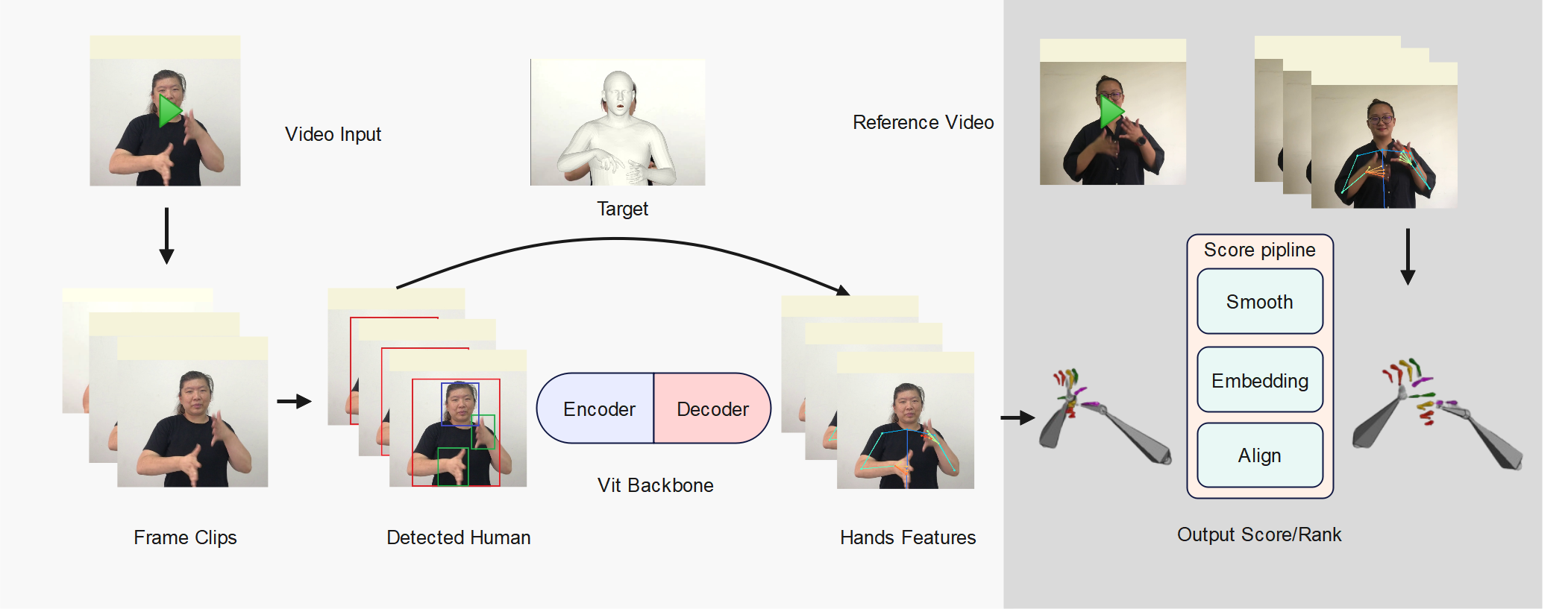} 
\caption{The pipeline of our system.} 
\label{Fig.1} 
\end{figure*}

\section{related work}
\subsection{Sign Language Pose Reconstruction}
The recognition of hand postures is a hot research topic, and recent efforts have focused on directly recovering hand joints \cite{ref23,ref24,ref25}. However, sign language involves actions that include the face and torso, and focusing solely on the hands in teaching environments can create a sense of disconnection and unreality. The goal of 3D Whole-Body Mesh Recovery is to accurately capture the full human body's posture and shape, with methods based on smplx \cite{ref26} supporting the reconstruction of hand keypoints and meshes well, as well as being useful for generating skeletal animations. Early pose estimation methods either focused only on body movements, ignoring hand actions, or focused solely on hand postures without considering body posture. Shuai et al.'s EasyMocap \cite{ref27} performs well in recovering hands from RGB images, but its multi-view design is costly. Yu et al.'s motion capture system FrankMocap \cite{ref28} can capture both body and hand 3D postures in natural environments using monocular vision, but its real-time performance is poor. On the other hand, Zhe et al.'s real-time pose detection system OpenPose \cite{ref29} can detect keypoints for body, hands, and face, but its hand pose estimation is less effective. Additionally, Gyeongsik et al.'s 3D full-body mesh estimation method Hand4Whole \cite{ref30} focuses on improving hand posture estimation by using separate networks and specific fusion modules to enhance performance, but the independent estimation of body, hand, and face parameters leads to network inconsistency, resulting in less natural and realistic estimation results. Recent studies \cite{ref19} have adopted the Transformer architecture, which has been successful in other computer vision tasks, as an alternative to convolutional networks. Lin \cite{ref2} et al.'s research has built a one-stage mesh recovery algorithm on this basis, using component-aware techniques to regress the body, hands, and face separately, achieving the best results on the AGORA dataset. Subsequent work \cite{ref21}, \cite{ref22} has continued this architecture and achieved the best results on some human body reconstruction datasets.

\subsection{Action Similarity}

Previous studies on action similarity evaluation feedback have mostly focused on sports, medical, and ceremonial contexts, with little demand for hand and finger joint movements. Jain et al. \cite{ref36} proposed a method based on Siamese networks to evaluate the quality of performed actions. Long-fei et al. \cite{ref37} suggested modeling user behavior through head movements, gaze, hand movements, and touch. However, the hardware setup of Inertial Measurement Units (IMUs) and RGBD cameras is complex, making them difficult to deploy and use in everyday life. R. Morais et al. \cite{ref31} used methods such as Euclidean distance, Manhattan distance, and Chebyshev distance to calculate similarity, but this method requires high correspondence of movements and can result in significant deviations due to the student's body posture. Ye et al. \cite{ref32} proposed a method combining joint angles with Dynamic Time Warping (DTW) \cite{ref38} to calculate similarity, thereby improving the accuracy of similarity calculations. As research into similarity calculations has deepened, due to the influence of the human skeleton and the large computational load, the calculation of similarity has gradually shifted from considering distance features to considering vector and angle features. Fang Yu et al. \cite{ref33} introduced a method based on oriented joint vector descriptor features, assessing similarity by calculating vector cosine values, which can be used in virtual reality simulations. Shen et al. \cite{ref34} proposed a method using cascaded quaternions to calculate the distance of quaternions representing joint orientation, assessing human similarity. Xue et al. \cite{ref35} used DTW to compare test sequences with standard sequences, implementing similarity assessment functionality.

\begin{figure*}[h] 
\centering 
\includegraphics[width=0.9\textwidth]{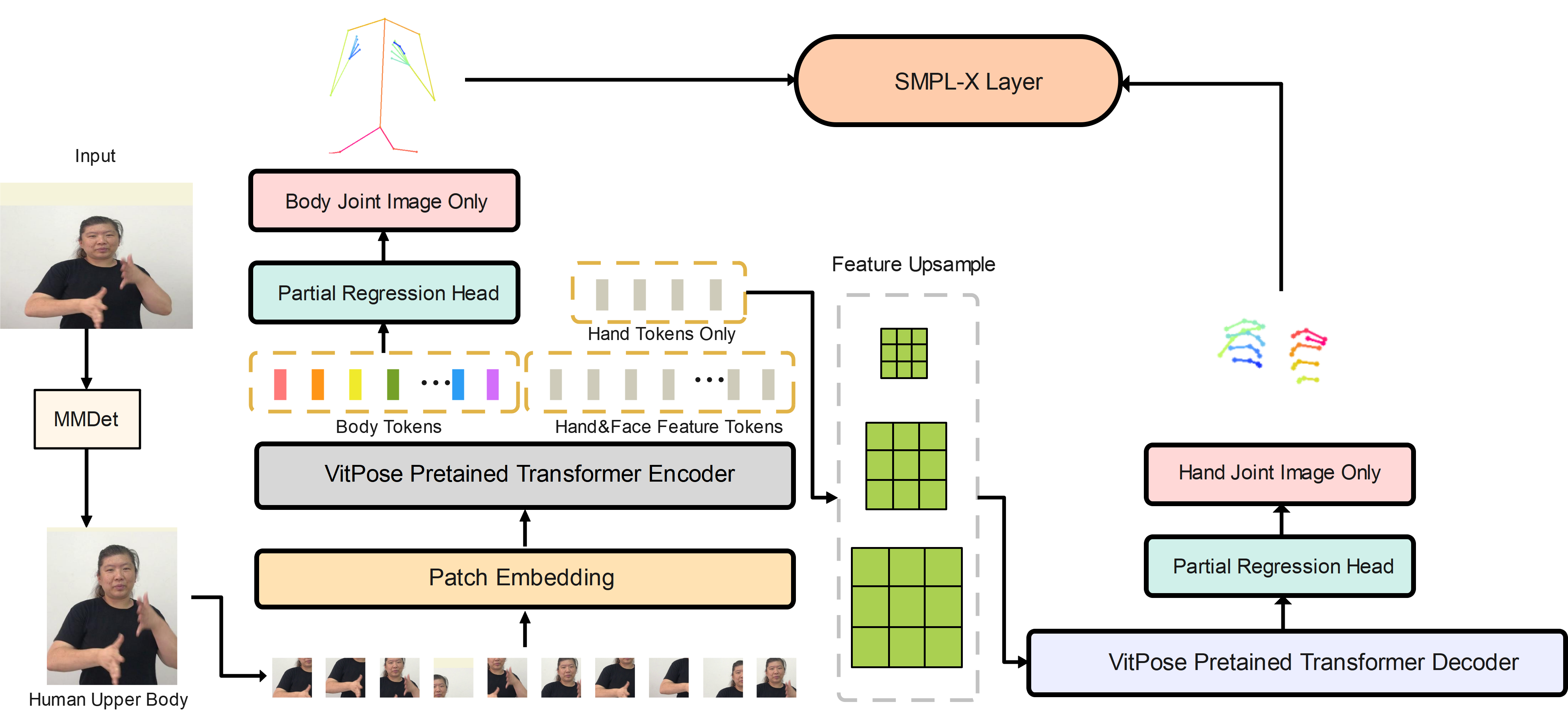} 
\caption{3D human mesh reconstruction} 
\label{Fig.2} 
\end{figure*}

\section{Two stage sign language scoring}
We will now describe in detail the input and output of the sign language evaluation pipeline, and the two stages of processing sign language, which is shown in \ref{Fig.1} Given the input video \( \mathbf{V} \in \mathbb{R}^{T\times H \times W \times 3 }\), where the \(t^{th}\) frame is an RGB image \( \mathbf{I}(t) \in \mathbb{R}^{H \times W \times 3 }\) with resolution \( H \times W\), our first-stage human mesh feature model (denoted as \( \mathcal{F} \)) directly outputs a set of parameters consisting of the 3D positions of \(N \times 2\) skeletal keypoints in the world coordinate system,\( \mathbf{M}\in\mathbb{R}^{ 2N \times 3}\).
\begin{align}
    \mathbf{M}(t)=\mathcal{F}(\mathbf{I}(t))
\end{align}

The model estimates frames of the video independently of their sequence. Next, the sequence \(\mathbf{M} \in \mathbb{R}^{T \times N \times 3} \) is passed to the post-processing regression model (denoted as \( \mathcal{D} \)) which completes the second stage of score computation. Specifically, SmoothNet\cite{ref7} filters out noise from the captured feature sequences in the temporal dimension, yielding a smoothed motion representation sequence \(\mathbf{S}=\mathrm{SmoothNet}(M)\) and the differences before and after smoothing \(\mathbf{C_l}\). The quaternion embedding module retrieves the representation \(\mathbf{Q}(t)\) of the hand joint positions for each frame. These are calculated concurrently during embedding and saved as \(\mathbf{C_e}\). The differentiable DWT module can use any calculated distance \(d(t_1,t_2)\) to measure the differences between two action keyframes and perform time-corresponding alignment, calculating distance differences \(\mathbf{C_a}\) and frame correlations \(\mathbf{R}\). The specific losses from the three steps are concatenated and fed into a fully connected layer for learning, yielding a final score \(\mathbf{O} \in \mathbb{R}^3 \).

Section 3.1 will detail the architecture of our first-stage human feature model, 3.2 explains the quaternion embedding, SmoothNet smoothing, and DWT alignment, section 3.3 completes the final score computation, and 3.4 discusses the training process.

\subsection{Hand Feature Recovery Model}
Recent studies\cite{ref2,ref3,ref4} have demonstrated that vision Transformers\cite{ref1} can replace traditional convolutional neural networks in 3D human reconstruction and pose estimation. A single-stage network using an encoder-decoder architecture can effectively recover expressive full-body mesh parameters, providing good feature representation and the advantage of parallel computation. The first stage will use this approach to extract the required hand features.

\noindent \textbf{Preprocessing} To obtain the most robust human expressions from diverse sign language environments, preprocessing of the raw graphics is necessary. Specifically, for the input frame \(\mathbf{I}(t) \in \mathbb{R}^{H \times W \times 3}\), from top to bottom, we need to identify the subject performing the sign language action and detect corresponding low-scale features such as the hands and face. We use MMdet object detection tools \cite{ref39} to predict the input human bbox from the raw image and crop it to get the human center \(\mathbf{H}(t) \in \mathbb{R}^{H' \times W' \times 3}\). It is worth noting that the latest methods\cite{ref4} propose integrating human detection into the encoder post-processing and have achieved SOTA results, but this coupled framework is not suitable for our task of single-person evaluation.

\noindent \textbf{VitPose-Backbone}

The input RGB frame \(I\) is encoded with a ViT backbone.As shown in Figure \ref{Fig.2}, we divide the input image into fixed-size image patches \(\mathbf{P}\in\mathbb{R}^{\frac{HW}{M^{2}}\times(M^{2}\times3)}\), where \(M\) is the image partition size. Dividing the image into patches improves computational efficiency and reduces computational pressure. The ViT model maintains a constant resolution, so each outputtoken spatially corresponds to a block in the input image.
Due to the lower resolution of the hand image patches, direct pose estimation might lead to insufficient accuracy. Therefore, we adopt a feature-level upsampling cropping strategy in the local decoder. Specifically, the feature token sequence \(\mathbf{T_{f}}^{\prime}\) obtained from the global encoder is reshaped into a feature map and upsampled through deconvolution layers into multiple higher resolution features \(\mathbf{T}_{hr}\) to obtain more detailed hand features. Moreover, the decoder also uses a keypoint-guided deformable attention mechanism for precise localization of hand keypoints:
{\small
\begin{align}CA(\mathbf{Q},\mathbf{V},p_q)=\sum_{l=1}^L\sum_{k=1}^KA_{lqk}W\mathbf{V}_l(\phi_l(p_q)+\Delta p_{lqk})
\end{align}
}

where \(l\) and \(k\) represent feature levels and keywords, \(A\) and \(W\) represent attention weights and learnable parameters. \(\phi(\cdot)\) and \(\Delta p\) are position adjustments and offsets. This attention mechanism focuses on local features around the hand keypoints, reducing unnecessary computations.

\begin{figure*}[htbp] 
\centering 
\includegraphics[width=0.75\textwidth]{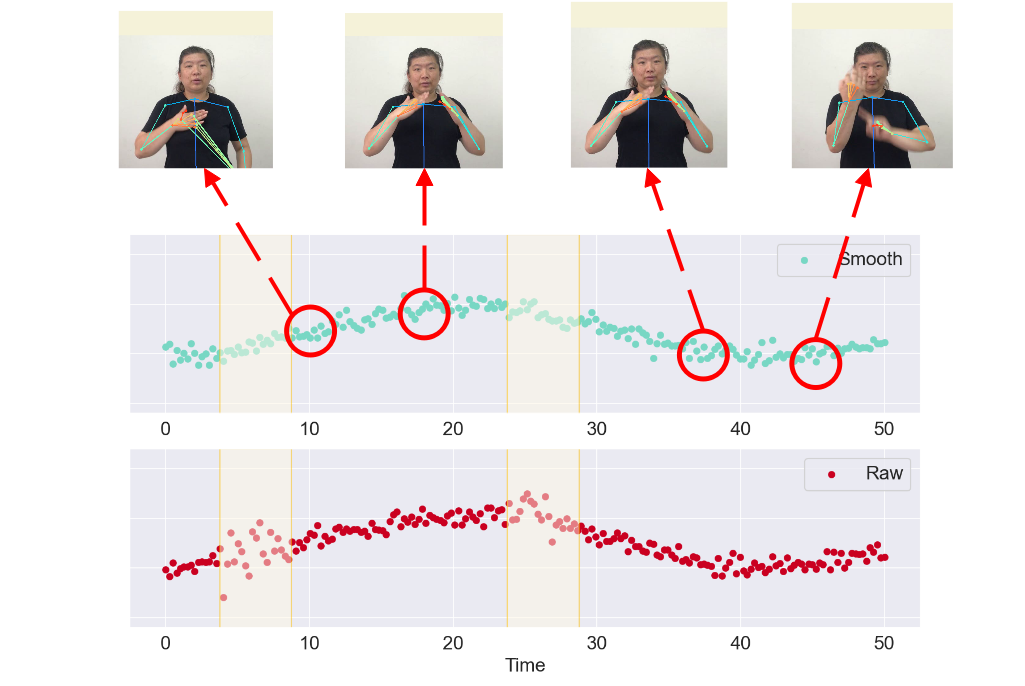} 
\caption{Smoothing pose estimation results} 
\label{Fig.3} 
\end{figure*}

\noindent \textbf{Body mesh representation}

Although our goal is hand feature estimation, the constraints of human dynamics can well limit the predicted model to be as physically accurate as possible according to physical conditions. We use SMPL-X\cite{ref5}, a model that includes hands and faces, to facilitate more accurate hand posture estimation.

The parametric 3D human model SMPL-X has input parameters for pose \(\theta \in \mathbb{R}^{53 \times3} \) , shape \(\beta \in \mathbb{R}^{10}\), and facial expressions \(\alpha \in \mathbb{R}^{10}\), outputting an expressive human-centered 3D mesh \( \mathbf{M} = \text{ SMPL-X}(\theta , \beta , \alpha ) \in \mathbb{R}^{V \times 3} \), where \( V = 10475 \) vertices. The mesh \( M \) is centered around the major keypoints, usually using the pelvic joint.

We use the mesh rather than directly regressing parameters from \(\mathbf{B}\) and \(\mathbf{L}\) based on specific considerations, which are discussed in the experiments. We merely treat mesh recovery as a task, using \(\mathbf{B}\) to regress body and camera parameters, and \(\mathbf{L}\) to regress hand and face parameters. We only use its hand parameters \(\mathbf{M}\) to complete the following tasks.

\subsection{Scoring Module}

The feature extraction module extracts hand parameters represented by axis-angle from each frame\(\mathbf{M}(t)\), which are obtained by sequentially rotating the human hand skeleton joints based on their parent-child relationships. Another set of motion poses \(\mathbf{M_s}(t)\), scored knownly, is simultaneously input as the basis for judging the quality of the actions.This is accomplished by comparing the similarity between the two sets of actions. Additionally, some necessary processing helps us better complete this process.

\noindent\textbf{Smoothing Motion}

A certain degree of jitter is observed in the prediction results. Besides systemic errors such as hand occlusions, this jitter could be due to individual differences in temporal motion, such as the speed of atomic motion transitions. To address this issue, we introduce SmoothNet\cite{ref7}, a data-driven, time-optimized general model for de-jittering, replacing traditional filters. SmoothNet inputs consecutive pose estimations and calculates position, velocity, and acceleration by examining changes in keypoint positions between adjacent frames, performing a linear fusion to output smoothed pose estimates, effectively improving result stability.

SmoothNet identifies errors that can arise from target loss due to mutual occlusion when estimating precise hand actions, as well as from non-smooth changes in true hand motion, which is considered undesirable to some extent as low proficiency or difficult-to-recognize dialects often produce such motion characteristics. The scores before and after smoothing can be used as a loss as follows:
\begin{align}
    C_s=\frac{|M(t)-\text{SmoothNet}(t)|}{|M(t)|}
\end{align}

\noindent\textbf{Quaternion-based Sign Language Action Embedding}

Researchers tend to use joint rotations rather than absolute coordinates to represent poses, reducing the impact of perspective and environmental factors. There are several forms of rotation representation, including direction cosine matrices, axis-angle, Euler angles, and quaternions. Considering computational speed and distance calculation comprehensively, we choose quaternions for their representation, which also avoids gimbal lock by using four degrees of freedom.

\begin{figure*}[h] 
\centering 
\includegraphics[width=0.65\textwidth]{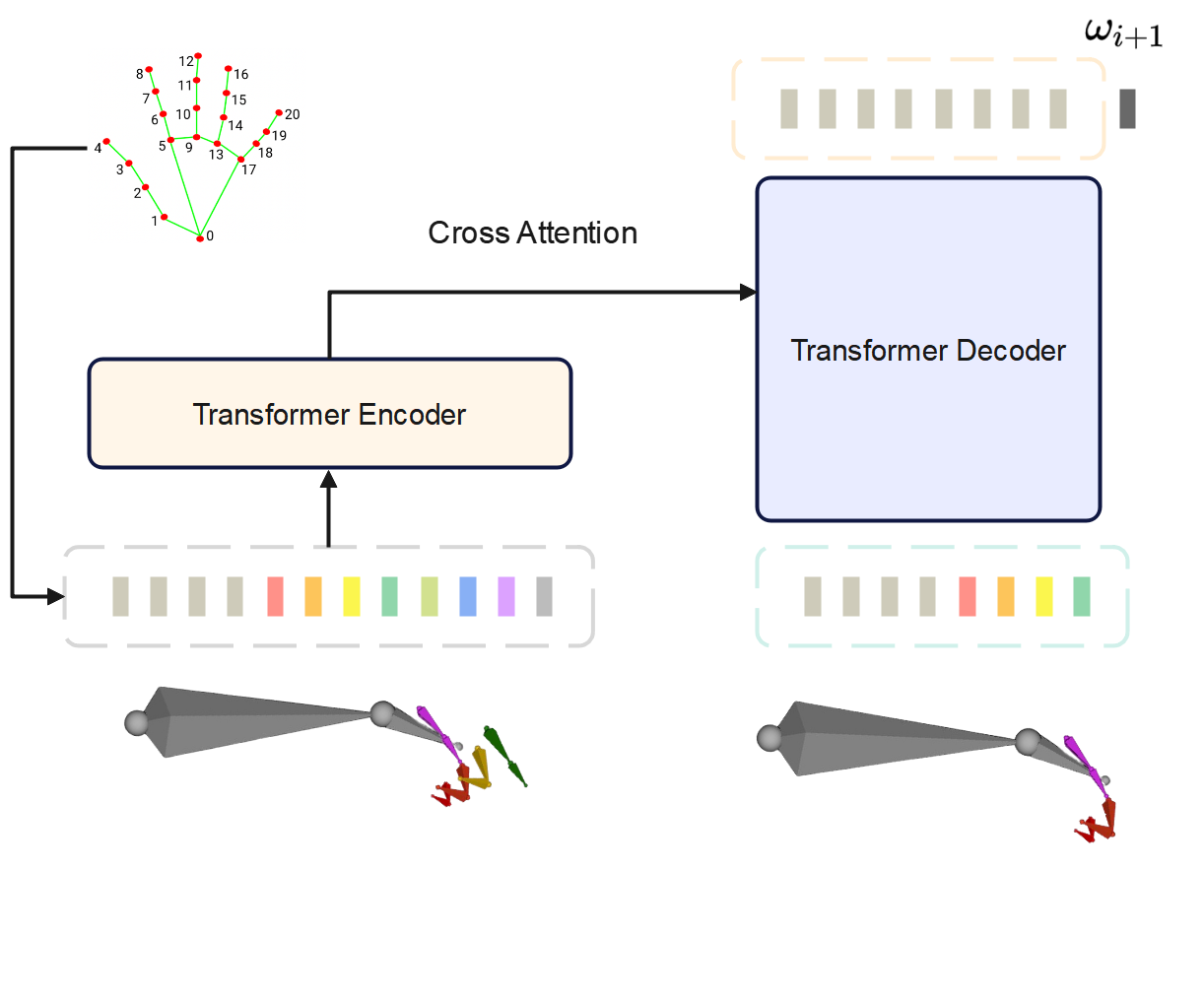} 
\caption{Embedding quaternion spatiotemporal sequences} 
\label{Fig.4} 
\end{figure*}

The human model consists of \(N\) joints, where the rotation of the \(i\)th joint can be represented by a quaternion \(q_i\). The position transformation of this joint relative to its parent joint can be calculated through the quaternion rotation formula:

\[ v'_i = q_i v_i q_i^{-1} \]

where \(v_i\) is the initial position vector of the \(i\)th joint in the coordinate system of the parent joint, \(v'_i\) is the position vector after rotation, \(q_i\) is the quaternion representing the rotation of the \(i\)th joint, and \(q_i^{-1}\) is the inverse quaternion of \(q_i\).

To represent the posture of the entire body, this transformation needs to be recursively applied, starting from the root joint. The global position of all joints is calculated through the local rotations of each joint:

\[ v'_{global,i} = \prod_{k=1}^{i} q_k v_i \left( \prod_{k=1}^{i} q_k
\right)^{-1} \]

Here, \(\prod_{k=1}^{i} q_k\) represents the cumulative product of all rotation quaternions from the root joint to the \(i\)th joint along the path, ensuring each joint's rotation considers its relative position and rotation within the entire chain.

The distance of the original action sequence can be calculated to judge motion differences, but the dimensions are not independent—one quaternion depends on another, and the difference in a parent joint directly affects the difference in a child joint, making it impossible to calculate vector distances directly. To address this, we use a weighted embedding approach to adjust and obtain low-dimensional expression differences, allowing each frame's motion to be treated as a vector for distance calculation.

The action sequence to be estimated, counted by frame \(t\), is defined as \(M(t)=v'_global(t)\), and at each specific time point \(t'\), this representation unfolds into a high-dimensional tensor \([p^1,p^2,...,p^N] \in \mathbb{R}^{4 \times N}\), each joint dimension \(p^i\) being recursively computed over its parent joint. A reference sequence can also be obtained in the same expression \(V_{s}(t)\). Using quaternion logarithmic differences as differences, where \(q^*\) is the quaternion operation representing the conjugate of \(q\):
\begin{align}
    df_i(t_1,t_2)=\log(q^i(t_1) \cdot (q_{s}^i(t_2))^*)
\end{align}

The \(N\) joints of the hand are calculated sequentially according to their parent-child relationship and can be considered as a sequence in spatial dimension. We introduce commonly used algorithms for processing time series to model and perform dimension embedding in space. The principle behind this approach is that the difference in child joints should always depend on the direction or difference of the parent joint. For a quaternion \(q_i(t)\), we can unfold it into a floating-point vector \(qv(t) \in \mathbb{R}^{4}\). Thus, the input can be viewed as a sequence with \(N\) time steps and 4 features, and we use the classic architecture of the Transformer\cite{ref6} to solve this translation-like problem. The encoder receives input \( \mathbf{R}^{N \times 4}\), undergoes a linear transformation to obtain \( \mathbf{K} \), \( \mathbf{Q} \), \( \mathbf{V} \), and uses self-attention mechanisms to obtain an embedded query sequence. These data will be used for attention queries on the standard sequence \(M_s\). The standard sequence input uses a masking mechanism, input sequentially according to the parent-child relationship, and can only observe information of the parent joint (only past information in terms of time). This will sequentially predict the error for each joint \( W = [w_1 ,w_2,\dots,w_N]\).

The fundamental reality of human motion suggests that excessive differences in parent joints render the differences in child joints almost meaningless, for example, an incorrect orientation at the base of a finger makes the differences at the fingertip negligible because the latter cannot compensate for the former but rather exacerbates the feedback mechanism. Using a learnable sequence is precisely to address this issue, and we employ a truncation strategy during inference, stopping the computation when the difference score \(w_i\) at a certain time step exceeds the threshold \(\mathcal{T}\), and directly adding a penalty term in the loss.

The aligned difference sequence is summed up to calculate the embedding distance between any two frames, where \(S\) is the calculation step reaching the threshold, and during training, \(S=N\).
\begin{align}
    D_i(t_1,t_2)=\sum^{S}_{i=0}w_i+\sum^{N}_{j=S+1}w_j
\end{align}

\noindent\textbf{Dynamic Time Warping Alignment}
Due to factors such as proficiency, intensity, and coherence of motion, which vary from person to person, these two sequences usually have different lengths in the time dimension. Conventional high-dimensional distances, such as Euclidean distance, cannot make accurate assessments. Therefore, we use derivative-based Dynamic Time Warping (DTW)\cite{ref8} to detect joints with larger gradients on the smoothed sequence.
\begin{align}
     DTW(Seq_1, Seq_2) = \min\left(\frac{\sqrt{\sum_{k=1}^{K} w_k}}{K}\right)
\end{align}

\begin{figure}[h] 
\centering 
\includegraphics[width=0.45\textwidth]{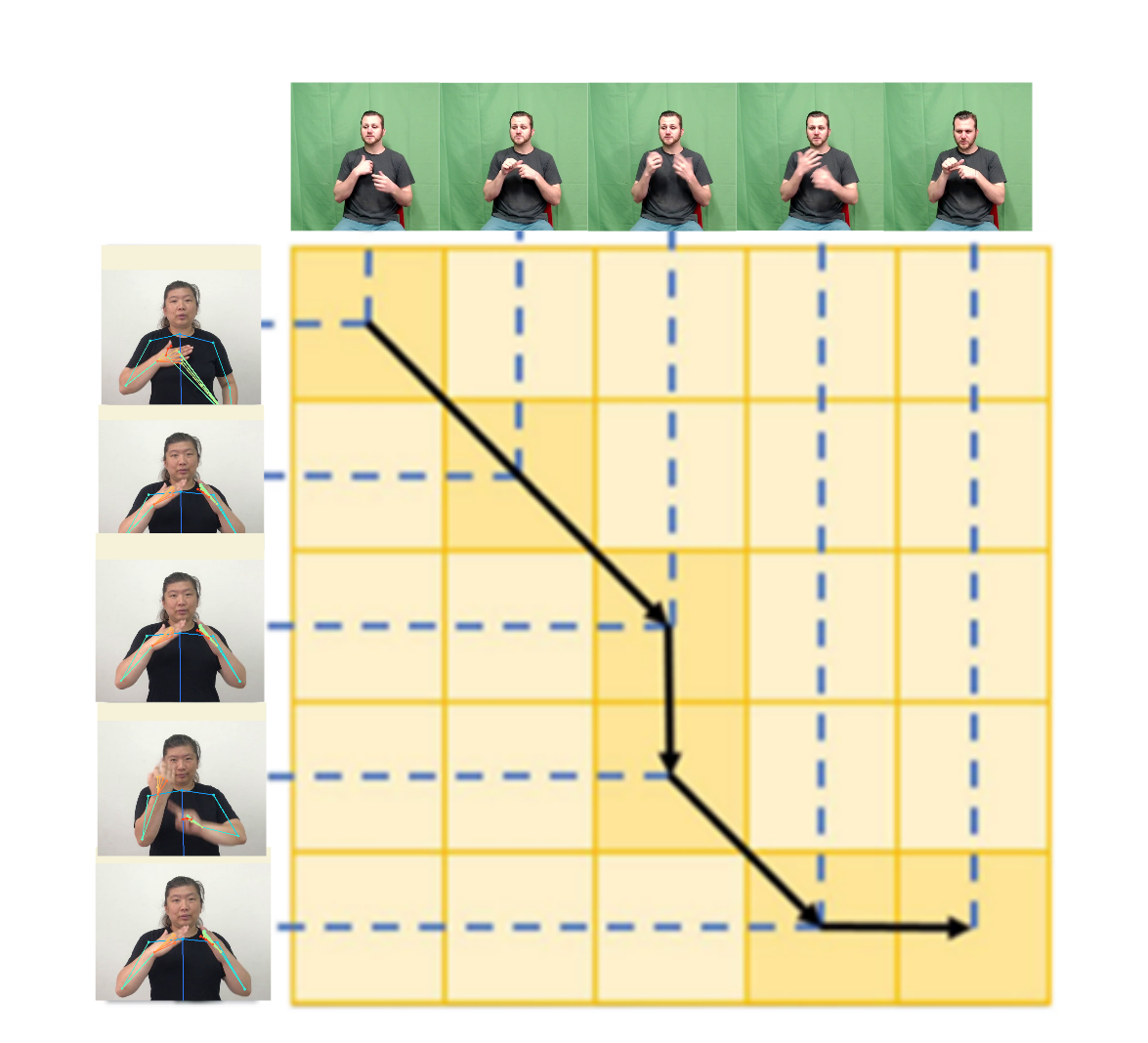} 
\caption{Align with Dtw } 
\label{Fig.5} 
\end{figure}
Quaternion interpolation can be performed between discrete frames to calculate the magnitude of the gradient, which as a measure of motion trends. It can to some extent eliminate differential features, focusing on the characteristics of the sign language itself. The adaptive joint weight distance of each motion sequence \(D(t)\) is taken in the time dimension for gradient \(\nabla_{\phi_i}(t)\) and processed with DTW computation.

\begin{align}
    D = DTW(\nabla_{\phi}(t),\nabla_{\phi_{s}}(t))
\end{align}

Time dimensions are aligned according to the shortest path for summation to obtain the difference between the two motion sequences, denoted as \( C_a \).

\subsection{Action Scoring}
The final score is derived from the outputs of the aforementioned three steps. We will concatenate the smoothing error \(C_s\) and the alignment error \(C_a\), and then feed them into a feedforward network, which will predict a three-dimensional assessment score \(\mathbf{O} \in \mathbb{R}^3 \)
\begin{eqnarray}
    \mathbf{O} = f(C_s \oplus  C_a)
\end{eqnarray}
These score dimensions can be trained according to downstream needs. In this study, we attribute its interpretability to the smoothness, completeness, and recognizability of the action, which are the three key features we believe the aforementioned framework extracts.
\begin{figure*}[h] 
\centering 
\includegraphics[width=0.8\textwidth]{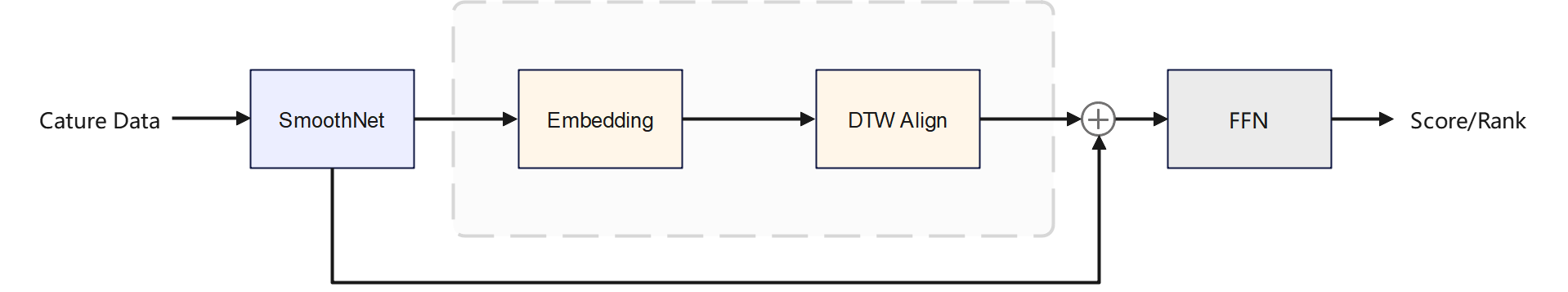} 
\caption{Scoring with FFN} 
\label{Fig.6} 
\end{figure*}

\subsection{Training Details}
Our system design is quite complex, involving multiple types of structures, so we have designed training tasks for each feature part, conducted asynchronously on different datasets, using differentiated losses.

\subsubsection{SCSL Dataset}

To support our experiments, we constructed a labeled Chinese Sign Language dataset (SCSL) based on a large collection of monocular sign language teaching videos provided by the Chinese Academy of Sciences. The dataset includes 600 complete teaching sentences performed by professional teachers, 468 clips collected from habitual sign language users, and 226 clips produced by sign language students. The latter two categories were scored by professional sign language teachers. Sign language sentences were manually divided into 6-8 sign language vocabulary segments, with their starting points serving as alignment markers as mentioned earlier, and each sign vocabulary also includes corresponding 3D sign language motion data segments. SCSL will be used for training the networks in both stages.

\subsubsection{Posture Network Training}
Our posture estimation training uses a vision Transformer model's backbone, fine-tuned on VitPose. It is trained through an end-to-end approach, with the task of reconstructing a complete SMPL-X human body model mesh and keypoint parameter.

\begin{eqnarray}
    L_\mathcal{F}=\lambda_1 L_{smplx}+ \lambda_2 L_{kpt3D}+ \lambda_3 L_{kpt2D}
\end{eqnarray}

This is essentially consistent with the methods\cite{ref4}, calculating these four components as L1 distances between the true values and the predicted values. However, we use some hyperparameters to fine-tune the penalties for prediction accuracy of each part, making the model more focused on the hand keypoints essential for our task.

Fine-tuning with the Transformer backbone provided by Vit-Pose, training on the Ubody\cite{ref2} dataset, and ultimately localizing on our dataset to obtain the first-stage estimation model.

\subsubsection{Embedding and Alignment Network Training}
We use scored video sequences, which can be viewed as RGB frames \(F(t)\), to learn embedding representations. To train accurate embeddings, we construct negative pairs from different actions and positive pairs from the same action but with different scores. To train alignment effectiveness, positive video pairs are annotated at certain checkpoints \(t^c_i\), where a baseline score \( s_{t^c_i}(I_1,I_2) \) based on the difference in annotated scores is given. We allow the baseline score to decrease along the time axis near these checkpoints, which can be accomplished using an exponential or Gaussian distribution \(g\).
Frames \(F(t_i) \longleftrightarrow  F_s(t_i)\) from the video pairs undergo Transformer computation to obtain a joint score sequence \(W(t)\), and these scores are directly computed as errors for each "spatial step." We use L1 errors to compute score differences and L2 errors to calculate differences between joints.
\begin{align}
    L_s=\sum^T_{t=0}|\sum^N_{n=0}w(n) - s_{t^c_i}g(t-t^c_i)|
\end{align}

\begin{align}
    L_t=\sum^T_{t=0}  \sum^N_{n=0} ||w(n)-d(j_n,j^s_n)||
\end{align}

\begin{align}
    L_\mathcal{D}= L_s + L_t 
\end{align}

Training takes place on our annotated dataset SCSL. SmoothNet is fixed at the input end of the model and does not participate in the training. We train the performance of embedding expression and alignment concurrently.

\subsubsection{Scoring Result Training}
Based on the sign language actions captured, analysis is performed to regress a score sequence. We use two parts of the loss to measure these predictions' differences. An L1 loss is introduced to measure the absolute distance from annotations \(L_{score}\), and another calculates a variance based on the ordinal ranking of scores, computed by correlation coefficients \(L_{rank}\).
\begin{align}
    L_\mathbf{S}=L_{score}+L_{rank}
\end{align}
The trained feedforward network should be able to fit the features obtained earlier into a percentile score with meaningful relative sizes. Also, this design allows for reasonable scoring with minimal tuning on this layer in a new scenario.

\section{Experiment}
\subsection{Posture Reconstruction Method Evaluation}

\textbf{Experimental Design}

To validate that the Hand Feature Recover Model in the first stage can provide more expressive features for the second stage, thereby improving the overall prediction accuracy of the evaluation algorithm, we prepared four groups of student learning videos with varying degrees of sign language proficiency. The groups are categorized as excellent (sign language scores between 90-100), good (80-90), average (70-80), and poor (60-70), each containing 20 videos. We implemented the first stage of the task using Easymocap, Expose, and Frankmocap, respectively, and input the extracted skeletal keypoint location parameters into the second stage for scoring prediction. We used t-tests to measure whether there are differences in sign language scores obtained using different recognition reconstruction methods in the first stage.

\textbf{Results}

The t-test results of Easymocap, Expose, Frankmocap, and ours compared to actual scores are shown in Table \ref{tab.1}.

\begin{table}[htbp]
  \centering
  \caption{T-test results}
  \begin{tabular}{lcccc}
    \toprule
    &{Easymocap} & {Expose} & {Frankmocap} & {ours} \\
    \midrule
    p-value & 0.016 & 0.034 & 0.056 & 0.871 \\
    Cohen's d & 1.043 & 1.964 & 0.572 & 0.229 \\
    \bottomrule
  \end{tabular}
  \label{tab.1} 
\end{table}

Used Easymocap, Expose, and Frankmocap to implement the first stage task, and t-tests were performed on the final scores corresponding to the three hand reconstruction methods and the true scores. The p-values for Frankmocap and our method are both greater than 0.05, which means that the null hypothesis is accepted, i.e., the final scores obtained using Frankmocap and our method do not significantly differ from the true scores. This indicates that the hand features extracted by these two methods can effectively distinguish the completion level of student sign language. Cohen's d measures the size of the difference between data sets. Compared to the other methods, our method has the smallest Cohen's d value, which means that the Hand Feature Recover Model in the first stage can provide effective features, thereby greatly reducing the difference between the scores predicted by the scoring module in the second stage and the true scores.

\subsection{Credibility Evaluation}
\textbf{Introduction of Datasets and Experimental Setups}

To verify the credibility of our evaluation algorithm, we conducted experiments on the SCSL dataset, which includes 226 videos of students learning sign language, each video at 65 frames per second, and about 300 frames per video. The true score of each sign language action video is the average of scores given by multiple professional sign language teachers.

\textbf{Results}

We evaluated the performance of our algorithm by calculating the Spearman's rank correlation coefficient between the scores predicted by the evaluation algorithm and the actual scores. The Spearman's rank correlation coefficient \( \rho \) is calculated by the formula
\begin{eqnarray}
\rho_{y,\hat{y}}=1-\frac{6\sum_{i=1}^Nd_i^2}{N(N^2-1)}
\end{eqnarray}

where \( y \) is the vector of actual scores, \( \hat{y} \) is the vector of predicted scores, and \( d_i \) is the rank difference between \( y_i \) and \( \hat{y_i} \) for the \(i\)-th video. A higher \( \rho \) value indicates better estimation performance. We used 180 videos as training samples and the rest as test samples. The average rank correlation coefficient of the test results after 200 training rounds was calculated to evaluate the performance of our algorithm.

\begin{figure}[htbp] 
\centering 
\includegraphics[width=0.5\textwidth]{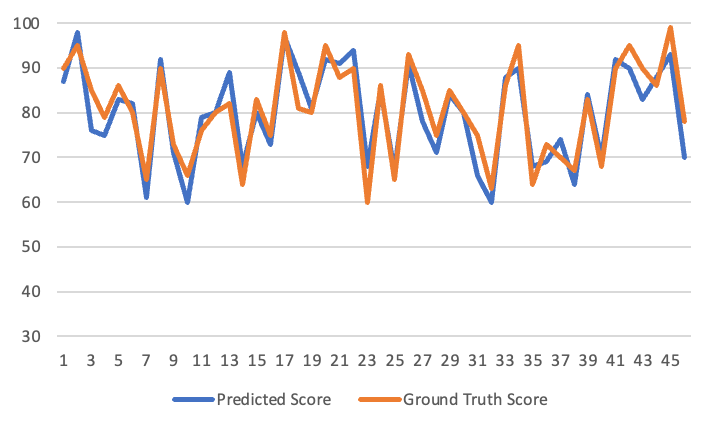} 
\caption{Predicted and actual score situation on the SCSL dataset} 
\label{Fig.8} 
\end{figure}

The computed results, as shown in Figure \ref{Fig.8}, indicate a Spearman correlation of 0.86 between the actual scores and the predicted scores. This suggests that our evaluation algorithm is highly consistent with the actual scoring of sign language actions, demonstrating the reliability and accuracy of the feedback provided to students by the platform. It can potentially replace teacher scoring, saving resources and improving the efficiency of teaching and learning.

\subsection{Algorithm Effectiveness Evaluation}
\textbf{Experimental Design}

To compare our proposed evaluation algorithm with other sign language assessment methods, we conducted experiments on the SHREC'17, DEVISIGN, and SCSL datasets, selecting 150 sign language videos from each dataset. We allowed other sign language assessment methods and our algorithm to score the selected videos. After scoring, all scores were categorized into tiers, with each tier spanning 5 points (95-100, 90-95, 85-90, etc.). If a video's predicted score from an algorithm falls within the same tier as its actual score, the algorithm is considered to have correctly predicted the score for that sign language video.

\textbf{Results}

\begin{table}[htbp]
\centering
\caption{Performance comparison on different datasets}
\label{tab:9}.
\begin{tabular}{llll}
\toprule
                                               & CSLR & DEVISIGN & SCSL \\
\midrule
CSITS                                          & 0.82     & 0.86     & 0.80 \\
SignInstructor                        & 0.77     & 0.81     & 0.83 \\
ours                                           & 0.85     & 0.86     & 0.95 \\
\bottomrule
\end{tabular}
\end{table}

As shown in Table \ref{tab:9}, for the CSLR dataset, our two-phase evaluation algorithm outperforms other assessment methods, with a prediction accuracy of 0.85. The DEVISIGN dataset, which contains videos of sign language vocabulary rather than longer sign language sentences, exhibits greater subjectivity in expert scoring of the completeness of signers' vocabulary in the videos, thereby increasing the discrepancy between predicted and actual scores. On the DEVISIGN dataset, our method and CSITS are tied for best performance, both with a prediction accuracy of 0.86. In our constructed SCSL dataset, which tests sign language sentence videos to minimize the impact of subjective factors and reduce variance in expert scoring, our method achieves the best prediction effect, with an accuracy of 0.95. This indicates that our two-phase sign language action evaluation algorithm can make full use of the expressive features provided in the first phase to achieve more accurate and effective sign language assessments.

\section{Conclusion and Future Work}

In this paper, we propose a two-stage sign language assessment pipeline that evaluates the quality of sign language actions directly from video comparisons. This method reconstructs the 3D representation of the human body from two-dimensional features in monocular RGB videos, uses quaternions to represent human joint rotations, and employs advanced sequential processing algorithms in space and time to extract differential features, ultimately regressing to quality rankings or scores depending on the application scenario. Our algorithm achieves high consistency with expert scoring and can partially replace professional sign language practitioners in scoring sign language , making it easily integrable into various digital sign language learning scenarios. Particularly, the pose intermediate representation we provide can offer seamless connectivity for systems that teach 3D sign language in virtual or augmented reality. There are still many areas for improvement in our algorithm. The target tasks for pose features extend beyond the scope of our research, and the complexity and real-time capabilities of the algorithm are future research directions. We will also work on annotating scores on more sign language datasets and refining our model. We hope this method can be applied to philanthropic causes to improve the learning and lives of people with disabilities.


\begin{thebibliography}{99}
\bibliographystyle{plain}

\bibitem{ref1} A. Dosovitskiy et al., “An Image is Worth 16x16 Words: Transformers for Image Recognition at Scale.” arXiv, Jun. 03, 2021. doi: 10.48550/arXiv.2010.11929.


\bibitem{ref2} J. Lin, A. Zeng, H. Wang, L. Zhang, and Y. Li, “One-Stage 3D Whole-Body Mesh Recovery With Component Aware Transformer,” presented at the Proceedings of the IEEE/CVF Conference on Computer Vision and Pattern Recognition, 2023, pp. 21159–21168. 

\bibitem{ref4} F. Baradel et al., “Multi-HMR: Multi-Person Whole-Body Human Mesh Recovery in a Single Shot.” arXiv, Feb. 22, 2024. Accessed: Apr. 08, 2024. [Online]. Available: http://arxiv.org/abs/2402.14654

\bibitem{ref3} Z. Cai et al., “SMPLer-X: Scaling Up Expressive Human Pose and Shape Estimation.” arXiv, Oct. 30, 2023. doi: 10.48550/arXiv.2309.17448.

\bibitem{ref6} A. Vaswani et al., “Attention is All you Need,” in Advances in Neural Information Processing Systems, Curran Associates, Inc., 2017. 

\bibitem{ref5} G. Pavlakos et al., “Expressive Body Capture: 3D Hands, Face, and Body From a Single Image,” presented at the Proceedings of the IEEE/CVF Conference on Computer Vision and Pattern Recognition, 2019, pp. 10975–10985. 

\bibitem{ref7} A. Zeng, L. Yang, X. Ju, J. Li, J. Wang, and Q. Xu, “SmoothNet: A Plug-and-Play Network for Refining Human Poses in Videos.” arXiv, Jul. 21, 2022. Accessed: Dec. 08, 2023. [Online]. Available: http://arxiv.org/abs/2112.13715

\bibitem{ref8} T. Górecki and M. Łuczak, “Multivariate time series classification with parametric derivative dynamic time warping,” Expert Systems with Applications, vol. 42, no. 5, pp. 2305–2312, Apr. 2015, doi: 10.1016/j.eswa.2014.11.007.

\bibitem{ref9} J. Joy, K. Balakrishnan, and S. M., “SiLearn: an intelligent sign vocabulary learning tool,” JET, vol. ahead-of-print, no. ahead-of-print, Aug. 2019, doi: 10.1108/JET-03-2019-0014.
\bibitem{ref10} F.-C. Yang, “Holographic Sign Language Interpreter: A User Interaction Study within Mixed Reality Classroom,” PhD Thesis, Purdue University, 2022.

\bibitem{ref11} L. Quandt, “Teaching ASL Signs using Signing Avatars and Immersive Learning in Virtual Reality,” in Proceedings of the 22nd International ACM SIGACCESS Conference on Computers and Accessibility, in ASSETS ’20. New York, NY, USA: Association for Computing Machinery, Oct. 2020, pp. 1–4. doi: 10.1145/3373625.3418042.

\bibitem{ref12} H.-Y. Li, Q. Lei, H.-B. Zhang, and J.-X. Du, “Skeleton Based Action Quality Assessment of Figure Skating Videos,” 2021 11th International Conference on Information Technology in Medicine and Education (ITME), pp. 196–200, Nov. 2021, doi: 10.1109/ITME53901.2021.00048.

\bibitem{ref13} P. Parmar and B. T. Morris, “Learning to Score Olympic Events,” in 2017 IEEE Conference on Computer Vision and Pattern Recognition Workshops (CVPRW), Honolulu, HI, USA: IEEE, Jul. 2017, pp. 76–84. doi: 10.1109/CVPRW.2017.16.

\bibitem{ref14} J. Xu, Y. Rao, X. Yu, G. Chen, J. Zhou, and J. Lu, “FineDiving: A Fine-grained Dataset for Procedure-aware Action Quality Assessment.” arXiv, Apr. 07, 2022. Accessed: Apr. 06, 2024. [Online]. Available: http://arxiv.org/abs/2204.03646

\bibitem{ref15} X. Feng, X. Lu, and X. Si, “Taijiquan Auxiliary Training and Scoring Based on Motion Capture Technology and DTW Algorithm,” International Journal of Ambient Computing and Intelligence, vol. 14, no. 1, doi: 10.4018/IJACI.330539.

\bibitem{ref16} H. Jain and G. Harit, “An Unsupervised Sequence-to-Sequence Autoencoder Based Human Action Scoring Model,” 2019 IEEE Global Conference on Signal and Information Processing (GlobalSIP), pp. 1–5, Nov. 2019, doi: 10.1109/GlobalSIP45357.2019.8969424.

\bibitem{ref17} Z. Liu, L. Pang, and X. Qi, “MEN: Mutual Enhancement Networks for Sign Language Recognition and Education,” IEEE Trans. Neural Netw. Learning Syst., vol. 35, no. 1, pp. 311–325, Jan. 2024, doi: 10.1109/TNNLS.2022.3174031.

\bibitem{ref18} Y. Zhang, Y. Min, and X. Chen, “Teaching Chinese Sign Language with a Smartphone,” Virtual Reality \& Intelligent Hardware, vol. 3, no. 3, pp. 248–260, Jun. 2021, doi: 10.1016/j.vrih.2021.05.004.

\bibitem{ref19} Y. Xu, J. Zhang, Q. Zhang, and D. Tao, “ViTPose: Simple Vision Transformer Baselines for Human Pose Estimation.” arXiv, Oct. 12, 2022. doi: 10.48550/arXiv.2204.12484.



\bibitem{ref21} Z. Cai et al., “SMPLer-X: Scaling Up Expressive Human Pose and Shape Estimation.” arXiv, Oct. 30, 2023. doi: 10.48550/arXiv.2309.17448.

\bibitem{ref22} F. Baradel et al., “Multi-HMR: Multi-Person Whole-Body Human Mesh Recovery in a Single Shot.” arXiv, Feb. 22, 2024. Accessed: Apr. 08, 2024. [Online]. Available: http://arxiv.org/abs/2402.14654

\bibitem{ref23} P. Panteleris, I. Oikonomidis, and A. Argyros, “Using a single RGB frame for real time 3D hand pose estimation in the wild,” in Proc. IEEE Winter Conf. Appl. Comput. Vis. (WACV), Mar. 2018, pp. 436–445.
\bibitem{ref24} C. Zimmermann and T. Brox, “Learning to estimate 3D hand pose from single RGB images,” in Proc. IEEE Int. Conf. Comput. Vis. (ICCV), Oct. 2017, pp. 4903–4911.
\bibitem{ref25} Z. Cao, G. Hidalgo, T. Simon, S.-E. Wei, and Y. Sheikh, “Realtime multi-person 2D pose estimation using part affinity fields,” in Proc. IEEE Conf. Comput. Vis. Pattern Recognit., Jul. 2017, pp. 7291–7299.


\bibitem{ref26} P. Panteleris, I. Oikonomidis, and A. Argyros, “Using a single RGB frame for real time 3D hand pose estimation in the wild,” in Proc. IEEE Winter Conf. Appl. Comput. Vis. (WACV), Mar. 2018, pp. 436–445.

\bibitem{ref27} C. Zimmermann and T. Brox, “Learning to estimate 3D hand pose from single RGB images,” in Proc. IEEE Int. Conf. Comput. Vis. (ICCV), Oct. 2017, pp. 4903–4911.

\bibitem{ref28} Z. Cao, G. Hidalgo, T. Simon, S.-E. Wei, and Y. Sheikh, “Realtime multi-person 2D pose estimation using part affinity fields,” in Proc. IEEE Conf. Comput. Vis. Pattern Recognit., Jul. 2017, pp. 7291–7299.


\bibitem{ref29} Z. Cao, G. Hidalgo, T. Simon, S.-E. Wei, and Y. Sheikh, “OpenPose: Realtime Multi-Person 2D Pose Estimation Using Part Affinity Fields,” IEEE Trans. Pattern Anal. Mach. Intell., vol. 43, no. 1, pp. 172–186, Jan. 2021, doi: 10.1109/TPAMI.2019.2929257.

\bibitem{ref30} G. Moon, H. Choi, and K. M. Lee, “Accurate 3D Hand Pose Estimation for Whole-Body 3D Human Mesh Estimation.” arXiv, Apr. 19, 2022. doi: 10.48550/arXiv.2011.11534.

\bibitem{ref31} R. Morais, V. Le, T. Tran, B. Saha, M. Mansour, and S. Venkatesh, “Learning Regularity in Skeleton Trajectories for Anomaly Detection in Videos,” presented at the Proceedings of the IEEE/CVF Conference on Computer Vision and Pattern Recognition, 2019, pp. 11996–12004. 

\bibitem{ref32} Y. Songtao and W. Xueqin, “Tai Chi video registration method based on joint angle and DTW [J],” Computing Technology and Automation, vol. 39, no. 1, pp. 117–122, 2020.

\bibitem{ref33} F. Yu, P. Jiazhen, and W. Jianhan, “The General Posture Descriptor of the Human Body for Virtual Reality Simulation,” in 2018 IEEE 9th International Conference on Software Engineering and Service Science (ICSESS), Nov. 2018, pp. 1147–1150. doi: 10.1109/ICSESS.2018.8663922.


\bibitem{ref34}J. Zhou et al., "Skeleton-based Human Keypoints Detection and Action Similarity Assessment for Fitness Assistance," 
2021 IEEE 6th International Conference on Signal and Image Processing (ICSIP), 2021, pp. 304-310, doi: 10.1109/ICSIP52628.2021.9689020.

\bibitem{ref35} J. Wang, C. Zeng, Z. Wang, and K. Jiang, “An improved smart key frame extraction algorithm for vehicle target recognition,” Computers \& Electrical Engineering, vol. 97, p. 107540, Jan. 2022, doi: 10.1016/j.compeleceng.2021.107540.

\bibitem{ref36} H. Jain, G. Harit, and A. Sharma, “Action quality assessment using Siamese network-based deep metric learning,” IEEE Trans. Circuits Syst. Video Technol., vol. 31, no. 6, pp. 2260–2273, Jun. 2021..
\bibitem{ref37} C. Long-fei, Y. Nakamura, and K. Kondo, “Modeling user behaviors in machine operation tasks for adaptive guidance,” 2020, arXiv:2003.03025.

\bibitem{ref38} M. Müller, Ed., “Dynamic Time Warping,” in Information Retrieval for Music and Motion, Berlin, Heidelberg: Springer, 2007, pp. 69–84. doi: 10.1007/978-3-540-74048-3\_4.


\bibitem{ref39} K. Chen et al., “MMDetection: Open MMLab Detection Toolbox and Benchmark.” arXiv, Jun. 17, 2019. doi: 10.48550/arXiv.1906.07155.

\end{thebibliography}
\end{document}